# Layer Pruning via Fusible Residual Convolutional Block for Deep Neural Networks

Pengtao Xu  
Peking University

Jian Cao  
Peking University

Fanhua Shang  
Xidian University

Wenyu Sun  
Peking University

Pu Li  
Peking University

## Abstract

*In order to deploy deep convolutional neural networks (CNNs) on resource-limited devices, many model pruning methods for filters and weights have been developed, while only a few to layer pruning. However, compared with filter pruning and weight pruning, the compact model obtained by layer pruning has less inference time and run-time memory usage when the same FLOPs and number of parameters are pruned because of less data moving in memory. In this paper, we propose a simple layer pruning method using fusible residual convolutional block (ResConv), which is implemented by inserting shortcut connection with a trainable information control parameter into a single convolutional layer. Using ResConv structures in training can improve network accuracy and train deep plain networks, and adds no additional computation during inference process because ResConv is fused to be an ordinary convolutional layer after training. For layer pruning, we convert convolutional layers of network into ResConv with a layer scaling factor. In the training process, the L1 regularization is adopted to make the scaling factors sparse, so that unimportant layers are automatically identified and then removed, resulting in a model of layer reduction. Our pruning method achieves excellent performance of compression and acceleration over the state-of-the-arts on different datasets, and needs no retraining in the case of low pruning rate. For example, with ResNet-110, we achieve a 65.5%-FLOPs reduction by removing 55.5% of the parameters, with only a small loss of 0.13% in top-1 accuracy on CIFAR-10.*

## 1. Introduction

Convolutional neural networks are extremely effective in a variety of computer visual tasks, such as image classification [18], object detection [7] and semantic segmentation [3]. Due to the development of computing power and the industrial demand on precision, many large network architectures with lots of layers have emerged, such as VGGNet [28], GoogleNet [32], ResNet [10] and MobileNet [13]. However, small scale models are required for deploying in embedded devices with weak computing power like mobile phones because of limited hardware resources. Currently, the methods for reducing the model size mainly include low-rank approximation [6], network quantization [5], network pruning [9].

Among them, network pruning is widely applied in various applications of engineering. The network pruning methods mainly include three types: weight pruning [9], filter pruning [24] and layer pruning [4]. Weight pruning is an unstructured pruning strategy, which requires special hardware structures and/or dedicated acceleration libraries to achieve speedup, while as structured pruning strategies, filter pruning and layer pruning can be deployed on any hardware device with no restrictions. However, when deploying neural networks on hardware such as GPUs and FPGAs, there is more data moving when model has more layers. So that in the case of the same pruning rate, more inference time and run-time memory usage is reduced by pruning layers than by pruning filters. However, there are very few studies on layer pruning. [4] pruned layers by training linear classifiers for each layer, but it is complicated to implement. Therefore, in this paper, we aim to obtain a simple layer pruning method, which can prune a large number of parameters and FLOPs while ensuring the accuracy of the model.

The goal of layer pruning is to remove unimportant layers with small loss of accuracy. To obtain the information of each layer's importance, we convert almost every convolutional layer of the original network into fusible residual convolutional block (ResConv), and introduce a layer scaling factor behind the batch normalization layer of ResConv. The ResConv structure is a residual block with one convolutional layer only, and we introduce a trainable infor-significance because some hardware is unable or difficult to deploy shortcut structures.

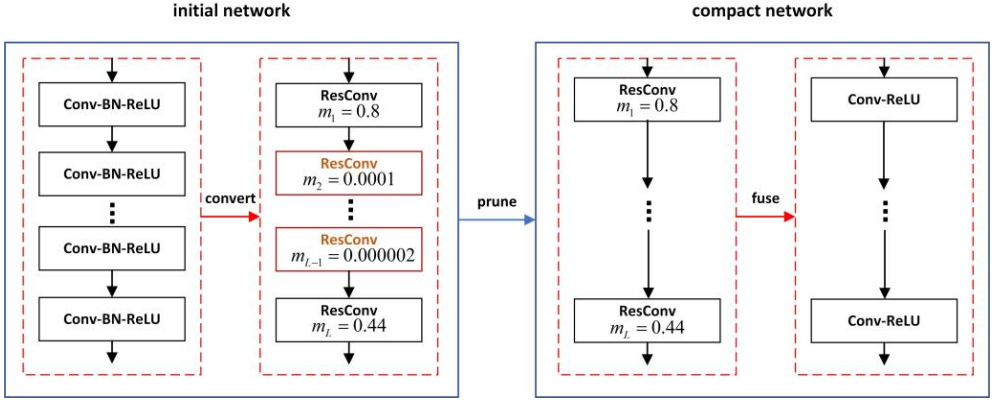

Figure 1. Framework of ResConv-Prune. In the unpruned network on the left, we first convert the Conv-BN-ReLU (convolution-batch normalization-relu) structure of the network into the fusible residual convolutional block (ResConv) with a layer scaling factor $m_i$ ($i$ means the $i$-th layer). Figure 2 below shows the ResConv, which is a residual structure with a trainable information control parameter. Then we make the factors of unimportant layers tend to zero through sparse training. Pruning these unimportant layers can obtain the pruned model on the right side, which can be improved by retraining to increase its accuracy. Finally, ResConv structures can be fused like Figure 3 below to get the final compact network.

mation control parameter in shortcut branch (see Figure 2 in Section 3.2). Unlike the building block in ResNet, our ResConv can be fused to be a convolutional layer (see Figure 3 in Section 3.2). The network containing ResConv is then sparsely trained by using L1 regularization to make the scaling factors tend to zero. After sparse training, we prune the layers with very small factors. When the pruning rate is relatively low, the pruned model can retain high accuracy without retraining, while when the pruning rate is high, the model needs to be retrained to ensure the accuracy. We call our pruning method ResConv-Prune, and Figure 1 shows its framework.

We conduct many experiments of different networks on the benchmark datasets, CIFAR-10 [17] and ImageNet [18]. The experimental results show that our pruning method only has a low accuracy loss, while a large number of parameters and FLOPs are pruned, and the overall performance exceeds the most advanced filter pruning [19, 12, 21, 14, 34, 22, 35, 20] and layer pruning [4] methods. Moreover, our experiments also verify that ResConv can be used to improve the accuracy of the model without any additional computation during inference process, and we find the positive role of ResConv for training deep plain networks without shortcut connections. It is of great

The main contributions of our work are summarized into threefold:

• Using the proposed ResConv structure, which can be fused to be an ordinary convolutional layer, can not only improve the accuracy of neural networks without increasing computation, but also train deep plain networks.

• The advantage of layer pruning over filter pruning is proved in terms of reducing inference time and run-time memory usage.

• Adequate experiments demonstrate the effectiveness of ResConv-Prune for model compression and acceleration. As a layer pruning method, ResConv-Prune achieves or even exceeds the performance of the most advanced filter pruning methods.

## 2. Related Work

In this section, we first introduce some works about shortcut connections. Then we revisit various pruning methods, including weight pruning, filter pruning and layer pruning.

**Shortcut Connections.** Theories about shortcut connections [1, 26, 33] have been studied for a long time. Some works such as [30, 31, 10] demonstrated that deep networks can be trained well by using shortcut connections.

In ResNet [10], it is common to use a shortcut structure in two or three convolutional layers. Unlike ResNet, we add a shortcut structure to each individual convolutional block. ResConv, though which is little worse than ResNet in performance of accuracy, can be fused into the form of ordinary convolutional layer, while the building block of ResNet cannot.

**Weight Pruning.** Han *et al*. [9] proposed to prune the unimportant connections with small weights in trained neural networks. [29] added gate variables to each weight, and then achieved higher compression rates by sparse training. Most of weights in the network after pruning by these methods are zero, so storage space can be significantly reduced. However, weight pruning methods [9, 8, 29, 2] can only achieve acceleration with special hardware platforms and software frameworks.

sparse training approach to obtain the information of each layer's importance.

**Layer Pruning.** Like filter pruning, layer pruning is also a structured pruning strategy. Currently, there are only few works about layer pruning. SSS [14] achieved pruning by introducing a scaling factor to the residual connection. However, this method actually prunes the block instead of the individual layer, and can only be used on networks with shortcut connections. Chen and Zhao [4] achieved the real layer pruning based on feature representation. This method trains linear classifiers for each layer to rank layers, which is complicated to implement. At the same time, [4] always needed to retrain the model after pruning, which is

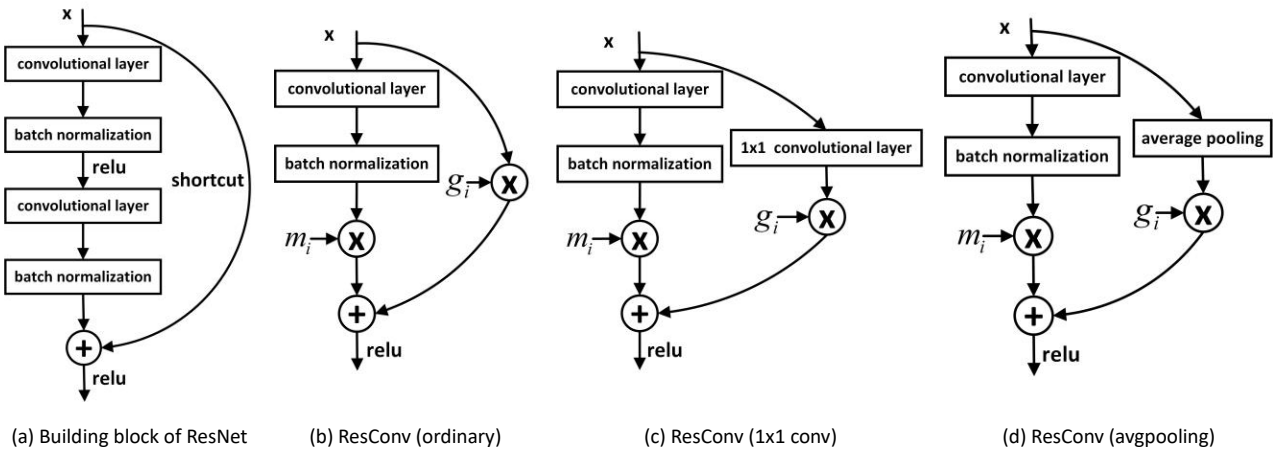

(a) Building block of ResNet  (b) ResConv (ordinary)  (c) ResConv (1x1 conv)  (d) ResConv (avgpooling)

Figure 2. The building block of ResNet and ResConv used in layer pruning. (a) shows the block used in ResNet. (b) illustrates the ordinary ResConv structure only with layer scaling factor $m_i$ and trainable information control parameter $g_i$. (c) shows the ResConv when the number of dimension varies, into which we introduce the $1 \times 1$ convolution with stride=1, padding=0. (d) shows the ResConv structure for downsampling, into which we introduce the average pooling with the same settings of stride, padding and kernel size as the convolutional layer of ResConv.

**Filter Pruning.** Unlike weight pruning, filter pruning removes the entire filters so there is no need for any special acceleration libraries. In [19], filters with small L1-norm are considered unimportant, so they can be pruned first. He *et al*. [11] achieved better pruning results using geometric median. Different from the above two methods based on property importance, Liu *et al*. [24] imposed L1 regularization on the scaling factors in batch normalization (BN) layers, and then selected unimportant filters to prune through sparse training. Our ResConv-Prune also uses this timeconsuming. Comparatively speaking, our method is easier to implement and requires no retraining when the pruning rate is low.

## 3. ResConv-Prune

In this section, we first discuss the advantages of layer pruning, so as to illustrate the necessity of studying layer pruning. Then we propose a novel ResConv structure. Finally, the adopted sparse training method and our pruning procedure are introduced.

## 3.1. Advantages of Layer Pruning

Like filter pruning, layer pruning is also a structured pruning method, and thus it can be applied to all hardware devices and does not require any dedicated acceleration library. However, filter pruning may make some layers only have very few channels or even one channel, which narrows the information transmission channel thus resulting bad model performance. In fact, the layer with few channels can be considered unimportant, which can be pruned completely by layer pruning but can not by filter pruning. Compared with filter pruning, another important advantage is that the pruned model obtained by layer pruning has less run-time memory usage and inference time because fewer layers means less data moving in memory.

## 3.2. Fusible Residual Convolutional Block

Simply speaking, layer pruning is to remove unimportant layers of the network, therefore, information about the importance of each layer is needed. In this paper, we convert the building block in ResNet that contains two or three convolutional layers into ResConv that contains only one convolutional layer (see Figure 2). By replacing the original convolutional structure with ResConv, and adding layer scaling factors, we can obtain the importance of each layer adaptively in training.

For many deep neural networks like ResNet [10] and MobileNetv2 [27] that have shortcut connections, too dimensions changes, we add a 1 × 1 convolution (see Figure 2(c)). The former two cases are based on the assumption that the sizes of feature maps do not change, while Figure 2(d) shows a different situation. When the shortcut go across feature maps of two sizes, we add an average pooling in the shortcut branch.

ResConv is only used in training. After training, it can be fused back to the original convolution structure (see Figure 3). In Figure 2, the batch normalization (BN) [15] is a linear operation as follows:

$$x_{bn} = \gamma \left( \frac{x - \mu}{\sigma} \right) + \beta \tag{1}$$

where $x$ and $x_{bn}$ denote the input and output of the BN layer, $\mu$ and $\sigma$ are the long term mean and standard deviations of the input $x$, respectively, and $\gamma$, $\beta$ are parameters to be learned. The BN layer can be fused into the convolutional layer [16]:

$$W_n = \frac{\gamma W_p}{\sigma} \tag{2}$$

$$b_n = \beta - \frac{\gamma(\mu - b_p)}{\sigma} \tag{3}$$

where $W_p$ and $b_p$ denote the weight and bias of the convolutional layer before BN layer, respectively, $W_n$ and $b_n$ are the weight and bias of the fused convolutional layer, respectively. Similarly, the layer scaling factors $m_i$ can also be fused into the convolutional layer. In other words, we many extra shortcut structures make the network difficult to train and degrade performance, and thus a trainable parameter $g_i$ is introduced to the shortcut branch to control the flow of information (see Figure 2).

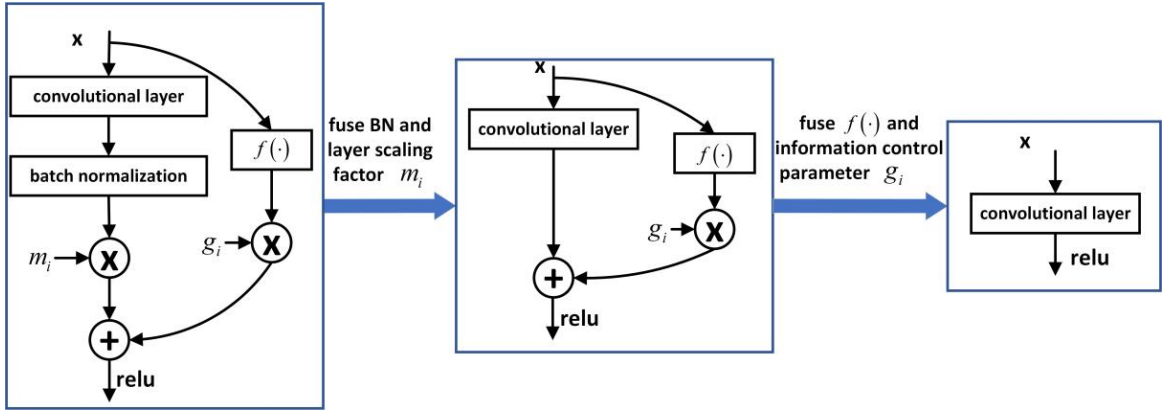

Figure 3. Fuse ResConv. $f(\cdot)$ is the operation performed on the shortcut branch.

As shown in Figure 2(b), the shortcut connections can be directly used in ResConv when the numbers of input and output channels are the same. When the number of

can ignore BN and $m_i$ of ResConv. Thus, ResConv can be represented as:

$$Y = X * W + b + g_i \cdot f(X) \tag{4}$$

where $*$ is the convolution operation, $X$ and $Y$ are the input and output feature maps of ResConv, respectively, $W$ and $b$

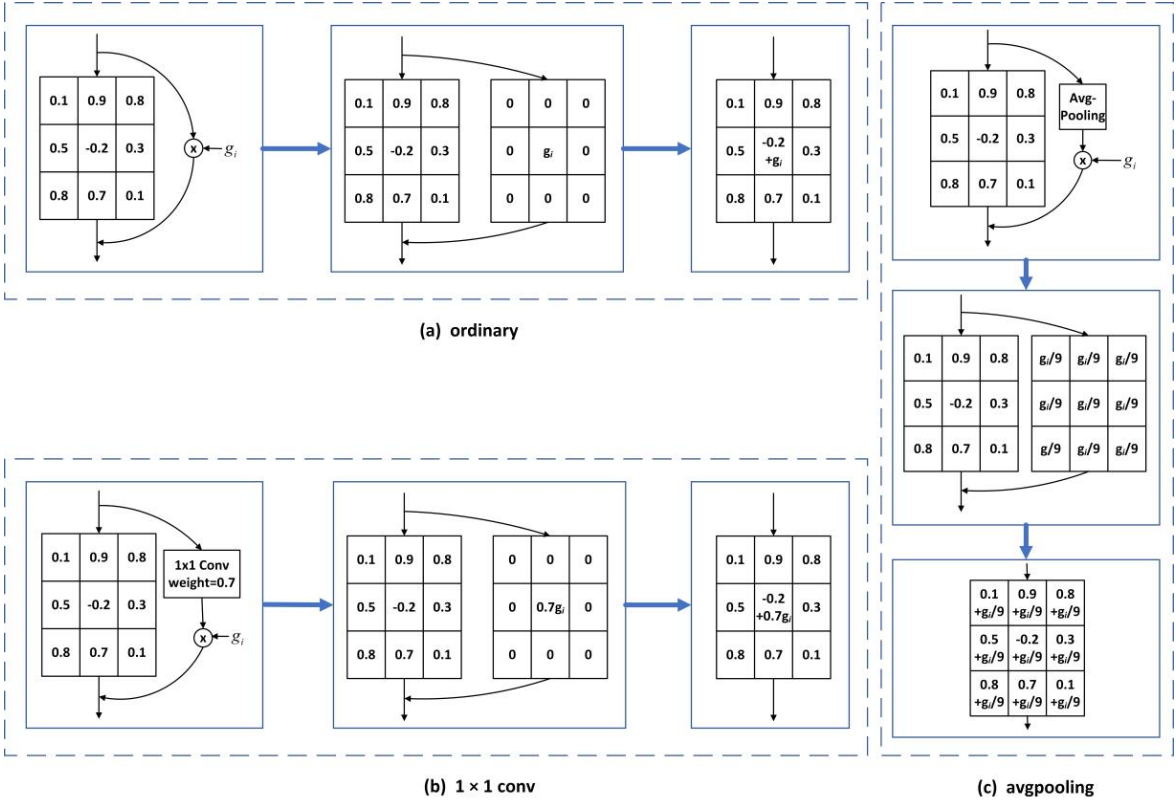

Figure 4. The process of fusion computation for ResConv. (a) denotes the fusion process when shortcut branch only has parameter $g_i$; (b) shows the fusion process when shortcut branch contains 1×1 convolution; (c) shows the fusion process when shortcut branch uses average pooling.

are the weight and bias of the convolutional layer, respectively, $g_i$ is the trainable information control parameter of shortcut connection, and $f(\cdot)$ is the operation performed on the shortcut branch. When the input and output feature maps have the same sizes and dimensions, $f(\cdot)$ conducts no operation, and $f(\cdot)$ is an average pooling operation, when ResConv does downsampling; when the number of dimensions changes between input and output, $f(\cdot)$ is the 1 × 1 convolution operation. Discrete convolution has the property of distributive law, so Eq.(4) can be regarded as an operation of one convolutional layer as follows:

$$Y = X * W_f + b_f \quad (5)$$

where $W_f$ and $b_f$ are the weight and bias of the convolutional layer which is the fusion of ResConv. $f(\cdot)$ in Eq.(4) can be turned into a convolution operation whose weight and bias are denoted by $W_s$ and $b_s$, so we can calculate $W_f$ and $b_f$ according to the following formulas:

$$W_f^{jq} = W^{jq} + g_i \cdot W_s^{jq}, \quad \forall j = 1, ..., t; q = 1, ..., u \quad (6)$$

$$b_f^{jq} = b^{jq} + g_i \cdot b_s^{jq}, \quad \forall j = 1, ..., t; q = 1, ..., u \quad (7)$$

where $j$ and $q$ represent the $j$-th output channel and the $q$-th input channel, respectively. For example, $W_s^{jq}$ means the values of $W_s$'s convolutional kernel in the the $j$-th output and the $q$-th input channel, and $t$ and $u$ are the total number of output and input channels, respectively. We can calculate $W_s$ by the following formulas (Here, we assume that the kernel size of the convolutional layer in ResConv is 3, and we can get similar formulas for other kernel sizes):

$$W_s^{jq} = \begin{cases} Padding(1), & j = q \\ Copy(0), & otherwise \end{cases} \quad (8)$$

$$W_s^{jq} = Padding(W_o^{jq}) \quad (9)$$

$$W_s^{jq} = \begin{cases} Copy(\frac{1}{3\times 3}), & j = q \\ Copy(0), & otherwise \end{cases} \quad (10)$$

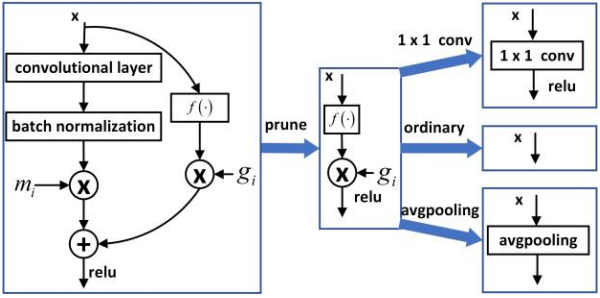

Figure 5. Prune layer in ResConv. We prune the layer with small $m_i$. And $g_i$ can be fused into last or next convolutional layer of the network. $f(\cdot)$ is the operation performed on the shortcut branch corresponding to Figure 2 (b), (c) and (d). In most cases, we prune the entire ResConv, while 1 × 1 convolution or average pooling is reserved in a few cases. However, the calculation of 1 × 1 convolution and average pooling is very small.

where $W_o$ is the weight of the 1×1 convolution on the shortcut branch, $Padding(\cdot)$ means padding zero around a single value to make it a 3 × 3 convolutional kernel, and $Copy(\cdot)$ means that a single value is copied 9 times to make a 3 × 3 convolutional kernel. Eqs.(8), (9) and (10) correspond to (b), (c) and (d) in Figure 2, respectively, and Figure 4 shows the process of fusion computation for ResConv. Besides, for (b) and (d) in Figure 2, $b_s$ is a zero vector, while for (c), $b_s$ is equal to $b_o$ which is the bias of the 1 × 1 convolution on the shortcut branch.

Finally, like the residual block in ResNet, ResConv can be used to train deep networks and improve model accuracy theoretically. In this paper, we set up some simple comparison experiments to verify this statement.

### 3.3. Sparse Training

When training the network weights, we need sparse training layer scaling factors at the same time to force them tend to zero. Then we can obtain the pruned model by removing the layers with small factors after training. There are many ways of sparse training, and we choose the most widely used L1 regularization, which is based on the L1norm. Therefore, the optimizing objective of the sparse training is given by:

$$Loss = \sum_{(x,y)} l(h(x,W),y) + \lambda \sum_{i=0}^{L} |m_i| \qquad (11)$$

where $x$ and $y$ denote the train input and target, $W$ is the trainable weights, $l(\cdot,\cdot)$ denotes the training loss of the network output and target, $|m_i|$ is the absolute value of the $i$-th layer scaling factor which acts as a sparsity-induced penalty, $L$ denotes the total number of layers in the network, and $\lambda$ is the sparsity factor which is used to balance two sum-terms.

### 3.4. Pruning Procedure

First, we turn the original network into one containing ResConv. The modified network is then sparsely trained. After training, we prune the layers with small layer scaling factors (see Figure 5). When the pruning rate is relatively low, there is no need to retrain the pruned network, while retraining is needed if the pruning rate is high. Finally, we can fuse the network to remove the ResConv structure from the network (see Figures 1 and 6).

## 4. Experiments

### 4.1. Datasets and Network Models

We conduct experiments on both the CIFAR-10 and ImageNet datasets to evaluate the effectiveness of our method.

**CIFAR-10.** The CIFAR-10 dataset contains 60,000 32 × 32 color images in 10 classes, including 50,000 training images and 10,000 test images. On the CIFAR-10 dataset, we experiment on the plain network VGG, the ResNet with shortcut structures, and the lightweight model MobileNet. The specific experimental networks are VGG16 with BN [19], ResNet-56/110 [10] and MobileNet for CIFAR-10 [23].

**ImageNet.** ImageNet is a larger dataset with a total of 1000 classes, containing 1.2 million training images and 50,000 validation images. For ImageNet, we conduct some experiments on ResNet-50 [10] and MobileNet [13].

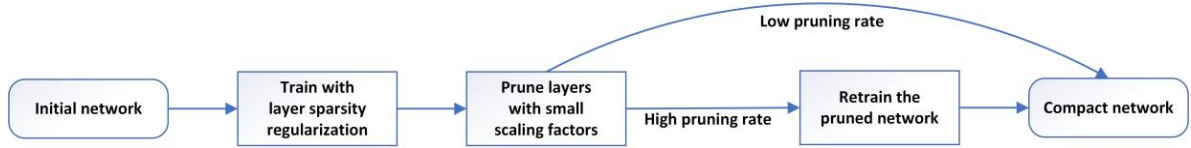

Figure 6. Pruning Procedure

## 4.2. Experimental Configurations

We use PyTorch [25] to implement our pruning method, and all experiments in this section are conducted on four NVIDIA GeForce RTX 2080Ti GPUs.

Table 1. The experimental results of VGG-16 on CIFAR-10. ResConv refers to converting convolutional layers of the original network into ResConv structures before training, and ResConvPrune represents our pruning method with retraining, while ResConv-Prune* means no retraining after pruning.

| Model | Top-1% | FLOPs(PR) | Parameters(PR) |
|---|---|---|---|
| VGG-16 | 94.15 | 314.29M(0.0%) | 14.99M(0.0%) |
| **ResConv** | 94.58 | 314.29M(0.0%) | 14.99M(0.0%) |
| **ResConv-Prune*($\lambda = 0.01$)** | 94.65 | 247.67M(21.2%) | 5.54M(63.0%) |
| SSS [14] | 93.21 | 183.69M(41.6%) | 3.94M(73.7%) |
| Zhao et al. [35] | 93.37 | 190.56M(39.3%) | 3.93M(73.8%) |
| **ResConv-Prune($\lambda = 0.1$)** | 93.45 | 138.53M(55.9%) | 3.61M(75.9%) |
| GAL [22] | 92.22 | 190.05M(39.5%) | 3.37M(77.5%) |
| Chen and Zhao [4] | 92.40 | 171.9M(45.3%) | 1.44M(90.4%) |
| HRank [20] | 92.53 | 109.17M(65.3%) | 2.65M(82.3%) |
| **ResConv-Prune($\lambda = 0.1$)** | 92.71 | 84.00M(73.3%) | 1.18M(92.1%) |
| GAL [22] | 90.92 | 172.45M(45.1%) | 2.68M(82.1%) |
| **ResConv-Prune($\lambda = 0.1$)** | 90.71 | 46.22M(85.3%) | 1.03M(93.1%) |

Table 2. The experimental results of ResNet-56 on CIFAR-10

| Model | Top-1% | FLOPs(PR) | Parameters(PR) |
|---|---|---|---|
| ResNet-56 | 93.69 | 126.55M(0.0%) | 0.85M(0.0%) |
| **ResConv** | 94.12 | 126.55M(0.0%) | 0.85M(0.0%) |
| L1 [19] | 93.49 | 91.96M(27.3%) | 0.73M(14.1%) |
| **ResConv-Prune*($\lambda = 0.001$)** | 93.75 | 81.00M(36.0%) | 0.61M(28.2%) |
| Chen and Zhao [4] | 92.19 | 75.7M(40.2%) | 0.42M(50.6%) |
| NISP [34] | 92.70 | 56.00M(55.7%) | 0.41M(51.8%) |
| **ResConv-Prune($\lambda = 0.01$)** | 92.72 | 46.37M(63.4%) | 0.35M(58.8%) |
| GAL [22] | 90.79 | 51.05M(59.7%) | 0.29M(65.9%) |
| HRank [20] | 91.15 | 33.58M(73.5%) | 0.27M(68.2%) |
| **ResConv-Prune($\lambda = 0.01$)** | 91.59 | 33.49M(73.5%) | 0.29M(65.9%) |

**Normal Training.** We use Stochastic Gradient Descent (SGD) with an initial learning rate of 0.1 to train networks. For CIFAR-10, the batch size, weight decay and momentum are set to 256, 0.0005 and 0.9. We totally train the network for 300 epochs with the learning rate being divided by 10 every 100 epochs. For ImageNet, the weight decay, momentum and number of training epochs are set to 0.0001, 0.9 and 120. And we set batch size to 512 for MobileNet, and 256 for ResNet-50.

**Sparse Training.** We initialize both the layer scaling factors $m_i$ and the trainable information control parameters $g_i$ to 1.0. For sparse training, there is a sparsity factor $\lambda$ selected between 0.0001 and 0.1 to control the sparsity degree. Other settings are the same as normal training.

**Pruning.** We need to define a pruning threshold to determine the pruning rate. The layers, whose scaling factors are less than the pruning threshold, will be pruned. In actual pruning, the prune threshold is determined by the desired pruning rate and the value of the sparsity factor.

Table 3. The experimental results of ResNet-110 on CIFAR-10

| Model | Top-1% | FLOPs(PR) | Parameters(PR) |
|---|---|---|---|
| ResNet-110 | 93.72 | 254.99M(0.0%) | 1.73M(0.0%) |
| **ResConv** | 94.15 | 254.99M(0.0%) | 1.73M(0.0%) |
| **ResConv-Prune*($\lambda = 0.0001$)** | 94.01 | 194.65M(23.7%) | 1.58M(8.7%) |
| HRank [20] | 93.58 | 107.80M(57.7%) | 0.71M(59.0%) |
| **ResConv-Prune*($\lambda = 0.001$)** | 93.59 | 88.06M(65.5%) | 0.77M(55.5%) |
| **ResConv-Prune($\lambda = 0.001$)** | 93.79 | 88.06M(65.5%) | 0.77M(55.5%) |
| GAL [22] | 92.77 | 132.30M(48.1%) | 0.96M(44.5%) |
| HRank [20] | 92.87 | 81.40M(68.1%) | 0.54M(68.8%) |
| **ResConv-Prune($\lambda = 0.01$)** | 93.67 | 82.19M(67.8%) | 0.53M(69.4%) |

Table 4. The experimental results of MobileNet on CIFAR-10

| Model | Top-1% | FLOPs(PR) | Parameters(PR) |
|---|---|---|---|
| MobileNet | 92.15 | 47.18M(0.0%) | 3.22M(0.0%) |
| **ResConv** | 93.11 | 47.18M(0.0%) | 3.22M(0.0%) |
| **ResConv-Prune*($\lambda = 0.001$)** | 93.03 | 29.77MM(36.9%) | 2.14M(33.5%) |
| **ResConv-Prune*($\lambda = 0.01$)** | 92.28 | 24.52M(48.0%) | 2.11M(34.5%) |
| **ResConv-Prune($\lambda = 0.1$)** | 90.46 | 16.23M(65.6%) | 1.78M(44.7%) |

**Retraining.** Since layer pruning may cause a large accuracy loss of the pruned network when pruning rate is high, retraining is needed to restore model accuracy. A small pruning rate has limited influence on the model accuracy, so retraining is not necessary. All the settings are the same as normal training.

## 4.3. Results and Analysis

In this subsection, we report the performance of our pruning method on reducing model parameters and FLOPs, and compare it with some state-of-the-art filter pruning and layer pruning methods, such as L1 [19], SSS [14], GAL [22], and HRank [20]. At the same time, we conduct some other experiments to verify the positive effect of ResConv structures to improve accuracy.

### 4.3.1 CIFAR10

Table 1 shows the results on VGG-16; Tables 2 and 3 show the results on ResNet-56/110; and Table 4 shows the results on the lightweight model, MobileNet.

**ResConv's role in improving model performance:** With application of ResConv in training, top-1 accuracy increases by 0.43% for both VGG-16 and ResNet-56/110, and by 0.96% for MobileNet. Although there is only a small improvement of performance, using ResConv adds no computation of the inference because ResConv can be fused after training. Therefore, using ResConv in training can be seen as an useful technique to improve the performance of the model.

**Results of pruning:** The experimental results show that our pruning method is extremely effective in reducing the model size and model computation. On VGG-16, we achieve 63.0% and 21.2% reduction in parameters and FLOPs, respectively, when getting 0.50% improvement in accuracy, without retraining. Compared with the advanced filter pruning method, HRank [20], we cut 73.3% (65.3% for HRank) of FLOPs and 92.1% (82.3% for HRank) of parameters with 0.18% higher accuracy. Similar results are achieved for ResNet-56/110. Even on the lightweight network, our ResConv-Prune performs well. On MobileNet, we prune 48.0% of FLOPs and 34.5% of parameters without retraining when the accuracy increases by 0.13%.

Table 5. The experimental results of ResNet-50 on ImageNet

| Model | Top-1% | Top-5% | FLOPs | Parameters |
|---|---|---|---|---|
| ResNet-50 | 76.15 | 92.87 | 4.11B | 25.56M |
| **ResConv** | **76.57** | **93.10** | **4.11B** | **25.56M** |
| **ResConv-Prune*($\lambda = 0.001$)** | 75.44 | 92.32 | 3.65B | 16.61M |
| GAL [22] | 71.95 | 90.94 | 2.35B | 21.26M |
| He et al. [12] | 72.30 | 90.80 | 2.75B | - |
| **ResConv-Prune*($\lambda = 0.01$)** | 73.88 | 91.30 | 2.17B | 13.63M |
| SSS [14] | 74.18 | 91.91 | 2.84B | 18.66M |
| HRank [20] | 74.98 | 92.33 | 2.32B | 16.21M |
| **ResConv-Prune($\lambda = 0.01$)** | 75.01 | 92.35 | 2.17B | 13.63M |
| GDP [21] | 69.58 | 90.14 | 1.59B | - |
| GAL [22] | 69.88 | 89.75 | 1.86B | 14.73M |
| **ResConv-Prune*($\lambda = 0.05$)** | 69.95 | 90.03 | 1.62B | 11.25M |

Table 6. The experimental results of MobileNet on ImageNet

| Model | Top-1% | Top-5% | FLOPs | Parameters |
|---|---|---|---|---|
| MobileNet | 70.87 | 89.93 | 578.83M | 4.23M |
| **ResConv** | **71.32** | **90.02** | **578.83M** | **4.23M** |
| **ResConv-Prune*($\lambda = 0.02$)** | 65.35 | 85.49 | 308.60M | 3.14M |

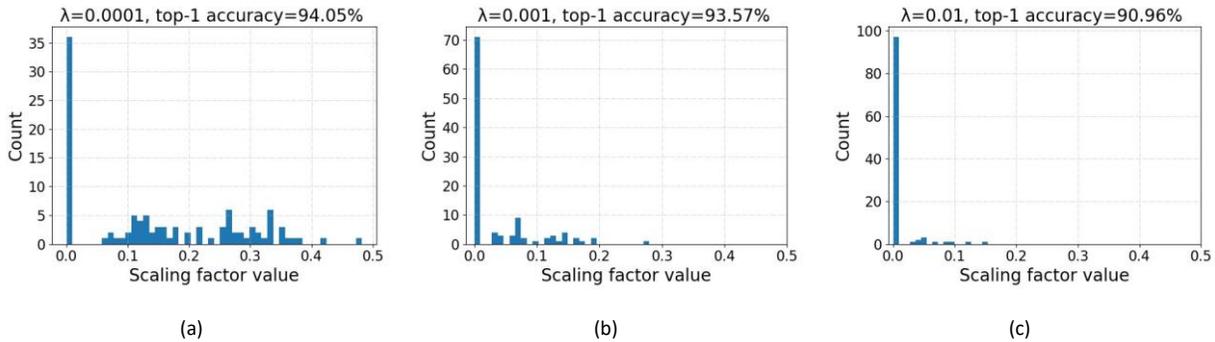

Figure 7. Distributions of layer scaling factors in a trained ResNet-110 under different sparsity factors $\lambda$. Scaling factors become sparser with the increase of $\lambda$.

#### 4.3.2 ImageNet

We conduct many experiments on ImageNet to validate the effectiveness of our approach on large-scale datasets. Table 5 shows the results on ResNet-50 and Table 6 shows the results on MobileNet. We find that the use of ResConv in training still improves the accuracy of the model, suggesting that ResConv can be an useful tool to improve accuracy. In addition, our method still achieves competitive results to the state-of-art filter pruning methods on ResNet-50, and can also prune layers successfully on MobileNet.

### 4.4. Other Experiments

#### 4.4.1 Time and Memory

In this subsection, we conduct experiments to explore the advantages of the layer pruning in terms of inference time and run-time memory usage.

We randomly generate layer pruning model and filter pruning model of VGG-16 with the same FLOPs and number of parameters. Then we use one GPU to test inference time (the time to infer the entire test set) and run-time memory usage of two models on the CIFAR-10 dataset. Table 7 shows that the layer pruning model takes less time and occupies less memory than the filter pruning model during inference process, which indicates that layer pruning is more advantageous than filter pruning.

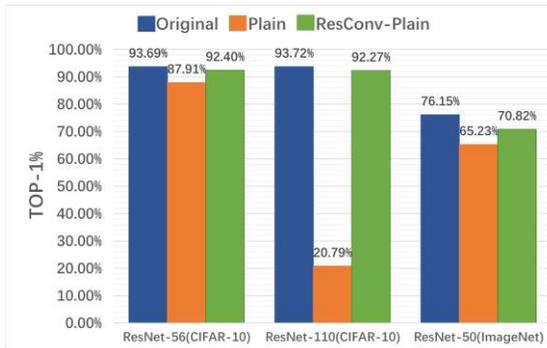

Figure 8. Accuracy comparison of original residual networks (Original), plain networks (Plain), and plain networks using ResConv when training (ResConv-Plain).

Table 7. Comparison of inference time (Time) and run-time memory usage (Memory) between filter pruning and layer pruning on CIFAR-10. (The results are the average of 100 experiments, and the batch size used is 64.)

| Method | FLOPs | Parameters | Time | Memory |
| --- | --- | --- | --- | --- |
| Filter Pruning | 68.5M | 4.2M | 2.01s | 1073MB |
| Layer Pruning | 68.5M | 4.2M | 1.65s | 1037MB |

4.4.2 Effect of Sparsity Factor

We can make networks sparser by using larger sparsity factor $\lambda$. In Figure 7, we show the distribution of layer scaling factors with different sparsity factor values. In this subsection, we present the experimental results by training ResNet-110 on CIFAR-10.

We can observe that the model trained with larger $\lambda$ has more scaling factors closing to zero, but has lower accuracy. Therefore, in practical applications, we need to trade off the pruning rate against the accuracy, and then choose the appropriate sparsity factor.

4.4.3 Training Deep Plain Network

In this subsection, we remove all the shortcut connections from the original ResNet to turn it into a plain network. Figure 8 shows the accuracy of the plain networks using and without using ResConv structures. It can be seen that on both CIFAR-10 and ImageNet, higher accuracy can be obtained by using ResConv structures during training. Although the accuracy is not as good as the original ResNet, ResConv can be fused at the end of training, resulting in a plain network without shortcut connections. In addition, Figure 8 shows that ResNet-110 can hardly be trained without shortcut connections, while the plain network with ResConv can be trained to get high accuracy. In other words, it is possible to train deep plain networks by using ResConv. Because some hardware devices are difficult to deploy shortcut structures, it is of great engineering significance to train deep plain networks.

## 5. Conclusions

In this paper, we proposed an efficient layer pruning method, called ResConv-Prune. It first replaces the convolutional layers in original network with Resconv, into which layer scaling factors are also introduced. Then the unimportant layers are automatically identified and pruned after sparse training. Finally, the ResConv that are not pruned will be fused to be ordinary convolutional layers. On both simple and complex datasets, our pruning method exceeds the performance of state-of-the-art filter pruning and layer pruning methods. In addition, our method needs no retaining when the pruning rate is relatively low. We also verified that layer pruning model has less inference time and run-time memory usage for the same pruning rate than filter pruning. Moreover, we verified that the ResConv structure can be used to improve network accuracy and train deep plain networks, which is very meaningful in engineering applications.

## References

[1] Christopher M Bishop et al. *Neural networks for pattern recognition*. Oxford university press, 1995. 2

[2] Miguel A Carreira-Perpinán and Yerlan Idelbayev. "learning-compression" algorithms for neural net pruning. In *Proceedings of the IEEE Conference on Computer Vision and Pattern Recognition*, pages 8532–8541, 2018. 2

[3] Liang-Chieh Chen, George Papandreou, Iasonas Kokkinos, Kevin Murphy, and Alan L Yuille. Deeplab: Semantic image segmentation with deep convolutional nets, atrous convolution, and fully connected crfs. *IEEE Transactions on Pattern Analysis and Machine Intelligence*, 40(4):834–848, 2017. 1

[4] Shi Chen and Qi Zhao. Shallowing deep networks: Layerwise pruning based on feature representations. *IEEE Transactions on Pattern Analysis and Machine Intelligence*, 41(12):3048–3056, 2018. 1, 2, 3, 6

[5] Wenlin Chen, James Wilson, Stephen Tyree, Kilian Weinberger, and Yixin Chen. Compressing neural networks with the hashing trick. In *International Conference on Machine Learning*, pages 2285–2294, 2015. 1

[6] Emily L Denton, Wojciech Zaremba, Joan Bruna, Yann LeCun, and Rob Fergus. Exploiting linear structure within convolutional networks for efficient evaluation. In *Advances*

*in Neural Information Processing Systems*, pages 1269–1277, 2014. 1

[7] Ross Girshick, Jeff Donahue, Trevor Darrell, and Jitendra Malik. Rich feature hierarchies for accurate object detection and semantic segmentation. In *Proceedings of the IEEE Conference on Computer Vision and Pattern Recognition*, pages 580–587, 2014. 1

[8] Song Han, Huizi Mao, and William J Dally. Deep compression: Compressing deep neural networks with pruning, trained quantization and huffman coding. *arXiv preprint arXiv:1510.00149*, 2015. 2

[9] Song Han, Jeff Pool, John Tran, and William Dally. Learning both weights and connections for efficient neural network. In *Advances in Neural Information Processing Systems*, pages 1135–1143, 2015. 1, 2

[10] Kaiming He, Xiangyu Zhang, Shaoqing Ren, and Jian Sun. Deep residual learning for image recognition. In *Proceedings of the IEEE Conference on Computer Vision and Pattern Recognition*, pages 770–778, 2016. 1, 2, 3, 6

[11] Yang He, Ping Liu, Ziwei Wang, Zhilan Hu, and Yi Yang. Filter pruning via geometric median for deep convolutional neural networks acceleration. In *Proceedings of the IEEE Conference on Computer Vision and Pattern Recognition*, pages 4340–4349, 2019. 3

[12] Yihui He, Xiangyu Zhang, and Jian Sun. Channel pruning for accelerating very deep neural networks. In *Proceedings of the IEEE International Conference on Computer Vision*, pages 1389–1397, 2017. 2, 7

[13] Andrew G Howard, Menglong Zhu, Bo Chen, Dmitry Kalenichenko, Weijun Wang, Tobias Weyand, Marco Andreetto, and Hartwig Adam. Mobilenets: Efficient convolutional neural networks for mobile vision applications. *arXiv preprint arXiv:1704.04861*, 2017. 1, 6

[14] Zehao Huang and Naiyan Wang. Data-driven sparse structure selection for deep neural networks. In *Proceedings of the European Conference on Computer Vision (ECCV)*, pages 304–320, 2018. 2, 3, 6, 7

[15] Sergey Ioffe and Christian Szegedy. Batch normalization: Accelerating deep network training by reducing internal covariate shift. *arXiv preprint arXiv:1502.03167*, 2015. 4

[16] Raghuraman Krishnamoorthi. Quantizing deep convolutional networks for efficient inference: A whitepaper. *arXiv preprint arXiv:1806.08342*, 2018. 4

[17] Alex Krizhevsky, Geoffrey Hinton, et al. Learning multiple layers of features from tiny images. 2009. 2

[18] Alex Krizhevsky, Ilya Sutskever, and Geoffrey E Hinton. Imagenet classification with deep convolutional neural networks. *Communications of the ACM*, 60(6):84–90, 2017. 1, 2

[19] Hao Li, Asim Kadav, Igor Durdanovic, Hanan Samet, and Hans Peter Graf. Pruning filters for efficient convnets. *arXiv preprint arXiv:1608.08710*, 2016. 2, 3, 6, 7

[20] Mingbao Lin, Rongrong Ji, Yan Wang, Yichen Zhang, Baochang Zhang, Yonghong Tian, and Ling Shao. Hrank: Filter pruning using high-rank feature map. In *Proceedings of the IEEE/CVF Conference on Computer Vision and Pattern Recognition*, pages 1529–1538, 2020. 2, 6, 7

[21] Shaohui Lin, Rongrong Ji, Yuchao Li, Yongjian Wu, Feiyue Huang, and Baochang Zhang. Accelerating convolutional networks via global & dynamic filter pruning. In *IJCAI*, pages 2425–2432, 2018. 2, 7

[22] Shaohui Lin, Rongrong Ji, Chenqian Yan, Baochang Zhang, Liujuan Cao, Qixiang Ye, Feiyue Huang, and David Doermann. Towards optimal structured cnn pruning via generative adversarial learning. In *Proceedings of the IEEE Conference on Computer Vision and Pattern Recognition*, pages 2790–2799, 2019. 2, 6, 7

[23] Kuang Liu. Train cifar10 with pytorch. https://github.com/kuangliu/pytorch-cifar. 2017. 6

[24] Zhuang Liu, Jianguo Li, Zhiqiang Shen, Gao Huang, Shoumeng Yan, and Changshui Zhang. Learning efficient convolutional networks through network slimming. In *Proceedings of the IEEE International Conference on Computer Vision*, pages 2736–2744, 2017. 1, 3

[25] Adam Paszke, Sam Gross, Soumith Chintala, Gregory Chanan, Edward Yang, Zachary DeVito, Zeming Lin, Alban Desmaison, Luca Antiga, and Adam Lerer. Automatic differentiation in pytorch. 2017. 6

[26] Brian D Ripley. *Pattern recognition and neural networks*. Cambridge university press, 2007. 2

[27] Mark Sandler, Andrew Howard, Menglong Zhu, Andrey Zhmoginov, and Liang-Chieh Chen. Mobilenetv2: Inverted residuals and linear bottlenecks. In *Proceedings of the IEEE Conference on Computer Vision and Pattern Recognition*, pages 4510–4520, 2018. 3

[28] Karen Simonyan and Andrew Zisserman. Very deep convolutional networks for large-scale image recognition. *arXiv preprint arXiv:1409.1556*, 2014. 1

[29] Suraj Srinivas, Akshayvarun Subramanya, and R Venkatesh Babu. Training sparse neural networks. In *Proceedings of the IEEE Conference on Computer Vision and Pattern Recognition Workshops*, pages 138–145, 2017. 2

[30] Rupesh Kumar Srivastava, Klaus Greff, and Jurgen Schmid-̈ huber. Highway networks. *arXiv preprint arXiv:1505.00387*, 2015. 2

[31] Rupesh K Srivastava, Klaus Greff, and Jurgen Schmidhuber.̈ Training very deep networks. In *Advances in Neural Information Processing Systems*, pages 2377–2385, 2015. 2

[32] Christian Szegedy, Wei Liu, Yangqing Jia, Pierre Sermanet, Scott Reed, Dragomir Anguelov, Dumitru Erhan, Vincent

Vanhoucke, and Andrew Rabinovich. Going deeper with convolutions. In *Proceedings of the IEEE Conference on Computer Vision and Pattern Recognition*, pages 1–9, 2015. 1

[33] William N Venables and Brian D Ripley. *Modern applied statistics with S-PLUS*. Springer Science & Business Media, 2013. 2

[34] Ruichi Yu, Ang Li, Chun-Fu Chen, Jui-Hsin Lai, Vlad I Morariu, Xintong Han, Mingfei Gao, Ching-Yung Lin, and Larry S Davis. Nisp: Pruning networks using neuron importance score propagation. In *Proceedings of the IEEE Conference on Computer Vision and Pattern Recognition*, pages 9194–9203, 2018. 2, 6

[35] Chenglong Zhao, Bingbing Ni, Jian Zhang, Qiwei Zhao, Wenjun Zhang, and Qi Tian. Variational convolutional neural network pruning. In *Proceedings of the IEEE Conference on Computer Vision and Pattern Recognition*, pages 2780–2789, 2019. 2, 6